\documentclass[11pt]{article}
\usepackage{fullpage}
\usepackage{times}
\usepackage{amsmath,amsthm,amsfonts}
\usepackage[usenames,dvipsnames]{color}
\usepackage{graphicx}
\usepackage{mathrsfs}
\usepackage{algorithm,algorithmic}
\usepackage{url}
\usepackage{hyperref}

\newcommand{\gnding}[2]
  {\ensuremath{\mathcal{G}^{#1}_{#2}}}

\newcommand{\newstuff}[1]
{#1}

\newcommand{\mess}[3]
  {\ensuremath{\mu_{#1\to #2}(#3)}}

\newcommand{\etal}
  {{\it et al.}}
 
\newcommand{\rv}[1]
 {\ensuremath{#1}}

\newcommand{\srv}[1]
 {\ensuremath{\mathbf{#1}}}

\newcommand{\lv}[1]
 {\ensuremath{\mathtt{#1}}}

\newcommand{\anyset}[1]
 {\ensuremath{\mathcal{#1}}}

\theoremstyle{definition}

\title{Lifted Graphical Models: A Survey}
\author{Lilyana Mihalkova \& Lise Getoor \\ University of Maryland College Park}
\begin{document}
\maketitle
\section{Motivation and Scope}
Multi-relational data, in which entities of different types engage in a rich set of relations, is ubiquitous in many domains of current interest. For example, in social network analysis the entities are individuals who relate to one another via friendships, family ties, or collaborations; in molecular biology, one is frequently interested in modeling how a set of chemical substances, the entities, interact with, inhibit, or catalyze one another; in web and social media applications, a set of users interact with each other and with a set of web pages or other online resources, which may themselves be related via hyperlinks; in natural language processing tasks, it is often necessary to reason about the relationships between documents, or words within a sentence or a document. By incorporating such relational information into learning and reasoning, rather than relying solely on entity-specific attributes, it is usually possible to achieve higher predictive accuracy for an unobserved entity attribute, e.g., \cite{sen:aimag08}. For example, by exploiting hyperlinks between web pages, one can improve categorization accuracy \cite{craven:ecml98}.  Developing algorithms and representations that can effectively deal with relational information is important also because in many cases it is necessary to predict \newstuff{the existence of a relation} between the entities. For example, in an online social network application, one may be interested in predicting friendship relations between people in order to suggest new friends to the users; in molecular biology domains, researchers may be interested in predicting how newly-developed substances interact.

Given the diversity of applications that involve learning from and reasoning about multi-relational information, it is not surprising that the field of statistical relational learning (SRL)\newstuff{\cite{srl04-workshop,srl06-workshop,srl09-workshop,getoor:book07,starai10-workshop}} has recently experienced significant growth.
This survey provides a detailed overview of developments in the field. We limit our discussion to representations that can be seen as defining a graphical model using a relational language, or alternatively as ``lifted'' analogs of graphical models. Although in this way we omit discussions of several important representations, such as stochastic logic programs \cite{muggleton:ilp96} and ProbLog \cite{deraedt:ijcai07}, which are based on imposing a probabilistic interpretation on logical reasoning, by limiting the scope of the survey we are able to provide a more focused and unified discussion of the representations that we do cover. For these and other models, we refer the reader to \cite{deraedt:kdd-explorations03}. Because of the great variety of existing SRL applications, we cannot possibly do justice to all of them; therefore, the focus is on representations and techniques, and applications are mentioned in passing where they help illustrate our point.

The survey is structured as follows. In Section~\ref{sect:prelim}, we define SRL and introduce preliminaries. In Section~\ref{sect:srlmodels}, we describe several recently introduced SRL representations that are based on lifting a graphical model. Our goal in this section is to establish a unified view on the available representations by defining a generic, or template, SRL model and discussing how particular models implement its various aspects. In this way, we establish not just criteria for comparisons of the models, but also a common framework in which to discuss inference (Section~\ref{sect:inference}), parameter learning (Section~\ref{sect:weight}), and structure learning (Section~\ref{sect:structure}) algorithms. 
\section{Preliminaries}
\label{sect:prelim}
\subsection{What is SRL?}
Statistical relational learning (SRL) studies knowledge representations and their accompanying learning and inference techniques that allow for efficient modeling and reasoning in noisy and uncertain multi-relational domains. \newstuff{In classical machine learning settings, the data consists of a single table of feature vectors, one for each entity in the data. A crucial assumption made is that the entities in the data represent {\it independent and identically distributed (IID)} samples from the general population. In contrast, multi-relational domains contain entities of potentially different types that engage in a variety of relations. Thus, a multi-relational domain can be seen as consisting of several tables: a set of attribute tables, one for each entity {\it type,} that contain feature-vector descriptions for each corresponding entity, and a set of relationship tables that establish relationships among two or more of the entities in the domain. As a consequence of the relationships among the entities, they are no longer independent, and the IID assumption is violated.} A further characteristic of multi-relational domains is that they are typically noisy or uncertain. For example, there frequently is uncertainty regarding the presence or absence of a relation between a particular pair of entities. 

To summarize, an effective SRL representation needs to support the following two essential aspects: a) it needs to provide a language for expressing dependencies between different types of entities and their diverse relations; and b) it needs to allow for probabilistic reasoning in a potentially noisy environment.  
\subsection{Background and Notation}
Here we establish the notation and terminology to be used in the rest of this survey. SRL draws on both probability theory and on logic programming, which sometimes use the same term to describe different concepts. For example, the word ``variable'' could mean random variable (RV), or a logical variable. To avoid confusion, we distinguish between different meanings using different fonts, as summarized in Table~\ref{tbl:notation}.
\begin{table}
\begin{center}
\begin{tabular}{|l|l|l|} \hline
{\bf Concept} & {\bf Representation} & {\bf Example} \\ \hline \hline
Random variable (RV) & Upper-case letters & \rv{X}, \rv{Y} \\ \hline
Set of RVs & Bold upper-case letters & \srv{X}, \srv{Y} \\ \hline
Value assigned to RV& Lower-case letters & \rv{x}, \rv{y} \\ \hline
Set of values assigned to RVs & Bold lower-case letters & \srv{x}, \srv{y} \\ \hline
Logical variable & Typewriter upper-case letters & \lv{X}, \lv{Y} \\ \hline
Entity/constant & Typewriter lower-case letters & \lv{x}, \lv{y} \\ \hline
Set of items other than RVs & Calligraphic upper-case letters & \anyset{X}, \anyset{Y} \\ \hline
\end{tabular}
\end{center}
\caption{Notation used throughout this survey}
\label{tbl:notation}
\end{table}
\subsubsection{Terminology of Relational Languages}
\label{sect:languageterminology}
This section provides an overview of several commonly-used relational languages with a focus on the aspects that are most important to the rest of our discussion.
\paragraph{First-order Logic}
First-order logic (FOL) provides a flexible and expressive language for describing typed objects and relations. FOL distinguishes among four kinds of symbols: constants, variables, predicates, and functions \cite{russell:book03}. \emph{Constants} describe the objects in the domain, and we will alternatively call them \emph{entities.} For example, in the notation of Table~\ref{tbl:notation}, \lv{x} and \lv{y} are entities.  Entities are typically typed. \emph{Logical variables} act as placeholders and allow for quantification, e.g., \lv{X} and \lv{Y}. \emph{Predicates} represent attributes or relationships and evaluate to {\tt true} or {\tt false}, e.g., \lv{Publication(paper, person)}, which establishes a relation between a paper and an author, and \lv{Category(paper, category)}, which provides the category of a paper, are predicates, and the strings in the parentheses specify the types of entities on which these predicates operate. \emph{Functions} evaluate to an entity in the domain, e.g., \lv{MotherOf}. We will adopt the convention that the names of predicates and functions will start with a capital letter. The number of arguments of a predicate or a function is called its arity. A \emph{term} is a constant, a variable, or a function on terms. A predicate applied to terms is called an \emph{atom,} e.g., \lv{Publication(X, Y)}. A positive \emph{literal} is an atom and a negative literal is a negated atom. A \emph{formula} consists of a set of positive or negative literals connected by conjunction ($\wedge$) or disjunction ($\vee$) operators, e.g., $\neg\lv{Friends(X, Y)} \vee \lv{Friends(Y, X)}$. The variables in formulas are quantified, either by an existential quantifier $(\exists)$ or by a universal quantifier $(\forall)$. Here we follow the typical assumption that when no quantifier is specified for a variable, $\forall$ is understood by default. 
A formula expressed as a disjunction with at most one positive literal is called a \emph{Horn clause}; if a Horn clause contains exactly one positive literal, then it is a \emph{definite clause.} The positive literal in a definite clause is called the \emph{head,} whereas the remaining literals constitute the \emph{body.} Definite clauses can alternatively be re-written as an implication as $\mathtt{Body} \Rightarrow \mathtt{Head}$.
 Terms, literals, and formulas are called \emph{grounded} if they contain no variables. Otherwise, they are \emph{ungrounded.} Grounding, also called \emph{instantiation,} is carried out by replacing variables with constants in all possible type-consistent ways. The set of all groundings of $f$ will be denoted with  $\anyset{G}_{f}^{\anyset{C}}$, where \anyset{C} is a set of constraints that specify which groundings are allowed. In general, if predicate and function arguments are typed, type constraints are present by default.

To connect this FOL terminology to probability theory, we note that when the value of a ground atom is governed by a random process, it becomes a random variable with values in the set $\{\lv{true} , \lv{false}\}$. For example, let \lv{x} and \lv{y} represent two entities, i.e., specific individuals; then the grounded atom \lv{Friends(x, y)} represents the assertion that \lv{x} and \lv{y} are friends. If we are given a probability distribution that governs the value of \lv{Friends(x, y)}, we can reason about it in the same way in which we reason about ordinary random variables. In addition, it is helpful to treat unground atoms as {\it parameterized} RVs \cite{poole:ijcai03}, in the sense that once their variables, or parameters, are replaced by constants, they become RVs. For example, if \lv{X} and \lv{Y} are logical variables, \lv{Friends(X, Y)} is a parameterized RV because once we ground it by replacing the parameters \lv{X} and \lv{Y} with actual entities, we obtain RVs. We will refer to parameterized RVs as par-RVs for short, e.g. \lv{Friends(X, Y)} is a par-RV.
\paragraph{Object-Oriented Representations} 
As an alternative to FOL, the attributes and relations of entities can be described using an object-oriented representation (OOR). Here again, \lv{x} and \lv{y} represent specific entities in the domain, whereas \lv{X} and \lv{Y} are variables, or entity placeholders. As in FOL, entities are typed. Attributes and relations are expressed using a notation analogous to that commonly used in object-oriented languages. For example, \lv{x.Category} refers to the category of paper \lv{x}, whereas \lv{x.Author} refers to its authors. Inverse relations are also allowed, e.g., $\lv{y.Author^{-1}}$ refers to the papers of which \lv{y} is an author. Using this notation, chains of relations can be conveniently specified, e.g. $\lv{x.Author.Author^{-1}.Category}$ gives the set of categories of all papers written by the authors of \lv{x}. Note that because the \lv{Author} relation is typically one-to-many, in general, \lv{x.Author} refers to a set of entities, i.e., the set of all authors of the paper. Because of this, object-oriented languages allow for aggregation functions, such as \lv{mean}, \lv{mode}, \lv{max}, or \lv{sum}. For example,  we can write  $\lv{mode(x.Author.Author^{-1}.Category)}$. As in FOL, OOR statements can be grounded, or instantiated, by replacing variables with entities from the domain. 
Analogous to FOL, we will view ungrounded relation/attribute chains, as well as aggregations thereof, as par-RVs.
\paragraph{Structured Query Language (SQL)} 
\newstuff{It is natural to manipulate relational data, which is often stored in a relational database, using SQL. Thus, not surprisingly, SQL has been used as a representation in some of the SRL models discussed in the survey. For self-sufficiency, we provide a brief overview.} The attributes and relations of objects can be viewed as defining a relational schema, in which an attribute table corresponds to each entity type and a relationship table corresponds to each relation type in which entities can engage. It is therefore natural to manipulate such data using SQL. Here we review the {\tt select} statement, which has been used to represent relational dependencies in SRL models. For our purposes, the most useful form of the {\tt select} statement is expressed as follows:
\begin{center}
\begin{small}
\begin{tabular}{p{6cm}}
\begin{verbatim}
SELECT <column names>
FROM <table names>
WHERE <selection constraints>
\end{verbatim}
\end{tabular}
\end{small}
\end{center}
\subsubsection{Terminology of Probabilistic Graphical Models}
\label{sect:graphicalmodels}
\begin{figure}[t]
\begin{center}
\includegraphics[scale=0.4]{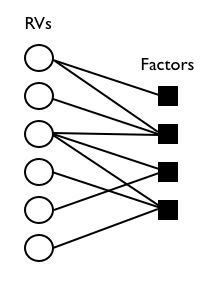}
\caption{Factor graph. Circular nodes correspond to variables, whereas square nodes correspond to factors. Variables are connected to the factors of which they are arguments. }
\label{fig:factorGraph}
\end{center}
\end{figure}

SRL also draws heavily on graphical models. Therefore, we next introduce basic concepts from that area. For a detailed introduction to graphical models, we refer the reader to \cite{koller:book09}. In general, to describe a probability distribution on $n$ binary RVs, one needs to store $2^n$ parameters, one for each possible configuration of value assignments to the RVs. However, frequently sets of RVs are conditionally independent of one another, and thus, many of the parameters will be repeated. To avoid such redundancy of representation, several graphical models have been developed that explicitly represent conditional independencies. One of the most general representations is the factor graph \cite{kschischang:infotheory01}. \emph{A factor graph} consists of the tuple $\langle\srv{X}, \anyset{F}\rangle$, where \srv{X} is a set of RVs, and \anyset{F} is a set of arbitrary but strictly positive functions called factors. It is typically drawn as a bipartite graph (Figure~\ref{fig:factorGraph}). The two partitions of vertices in the factor graph consist of the RVs in \srv{X} and the factors in \anyset{F} respectively. There is an edge between an RV \rv{X} and a factor $f$ if and only if \rv{X} is necessary for the computation of $f$; i.e., each factor is connected to its arguments. As a result, the structure of a factor graph defines conditional independencies between the variables. In particular, a variable is conditionally independent of all variables with which it does not share factors, given the variables with which it participates in common factors. 

A factor graph  $\langle\srv{X}, \anyset{F}\rangle$ defines a probability distribution over \srv{X} as follows. Let \srv{x} be a particular assignment of values to \srv{X}. Then,
\[P(\srv{X} = \srv{x}) = \frac{1}{Z} \prod_{f \in \anyset{F}} f(\srv{x}_f). \]
Above, $\srv{x}_f$ represents the values of those variables in \srv{X} that are necessary for computing $f$'s value. $Z$ is a normalizing constant that sums over all possible value assignments \srv{y} to \srv{X}, and is given by:
\begin{align}
Z = \sum_{\srv{y}}  \prod_{f \in \anyset{F}} f(\srv{y}_f). \label{eqn:factor}
\end{align}
\newstuff{As before, $\srv{y}_f$ represents the values of only those variables in \srv{Y} that are necessary to compute $f$.}

Factor graphs generalize two very common graphical models--Bayesian and Markov networks. A Bayesian network \cite{pearl:book88} is represented as a directed acyclic graph, whose vertices are the RVs in \srv{X}. The probability distribution over \srv{X} is specified by providing the conditional probability distribution for each node given the values of its parents. The simplest way of expressing these conditional probabilities is via conditional probability tables (CPTs), which list the probability associated with each configuration of values to the nodes.
A Bayesian network can be converted to a factor graph in a straightforward way as follows. For each node \rv{X}, we introduce a factor $\rv{f_X}$ to represent the conditional probability distribution of \rv{X} given its parents. Thus, $\rv{f_X}$ is computed as a function of only \rv{X} and its parents. 
In this case, the product is automatically normalized, i.e., the normalization constant $Z$ sums to 1.

A Markov network \cite{pearl:book88} is an undirected graphical model whose nodes correspond to the variables in \srv{X}. It computes the probability distribution over \srv{X} as a product of strictly positive potential functions defined over cliques in the graph, i.e. for any set of variables that are connected in a maximal clique, there is a potential function that takes them as arguments. A convenient representation for potential functions is the log-linear model, in which each potential function $\phi(\rv{X}_1 \dots \rv{X}_n)$ that is computed as a function of $n$ variables $\rv{X}_1 \dots \rv{X}_n$ is represented as the exponentiated product $\exp (\lambda \cdot f(\rv{X}_1 \dots \rv{X}_n))$. In this expression, $\lambda \in \mathscr{R}$ is a learnable parameter, and $f$ is a feature that captures characteristics of the variables and \newstuff{can evaluate to any value in $\mathscr{R}$}. In general, there may be more than one potential function defined over a clique. In this way,  a variety of feature functions, each with its own learnable parameter $\lambda$, can be defined for the same set of variables. Markov networks map directly to factor graphs---to convert a Markov network to a factor graph, for each maximal clique in the Markov network, we include a factor that evaluates to the product of potentials defined over that clique.

The advantage of discussing factor graphs rather than Bayesian networks, Markov networks, and others is that by describing algorithms for factor graphs, we make them immediately available to representations that can be viewed as specializations of factor graphs. This is especially true of inference algorithms. On the other hand, it will be beneficial to discuss at least some aspects of learning techniques separately for directed and undirected models. 
\section{Overview of  SRL models}
\label{sect:srlmodels}
Existing SRL representations can be split into two major groups. The first group consists of ``lifted'' graphical models -- representations that use a structured language to define a probabilistic graphical model. Representations in the second group impose a probabilistic interpretation on logical inference. As already discussed, to allow for greater depth, here we limit ourselves to the first group of languages. To provide a convenient representation that describes the common core of ``lifted'' graphical models, we start with par-factor graphs, short for parameterized factor graphs, defining them in the terminology of \cite{poole:ijcai03}. A par-factor graph is analogous to a factor graph \cite{kschischang:infotheory01} in that it generalizes a large class of SRL models and allows us, as much as possible, to present a unified treatment of them regardless of whether they are based on directed or undirected representations. 
\subsection{Par-Factor Graphs}
A par-factor graph is a ``lifted'' factor graph in the sense that, when instantiated, a par-factor graph defines a factor graph. It consists of a set of par-factors. Each par-factor is represented as a triple $(\anyset{A}, \phi, \anyset{C})$, where $\anyset{A}$ is a set of parameterized random variables, $\phi$ is a function that operates on these variables and evaluates to a strictly positive value, and $\anyset{C}$ is a set of constraints on how the variables may be instantiated. The par-factor graph is then just a set of par-factors 
$\anyset{F} = \{(\anyset{A}_i, \phi_i, \anyset{C}_i)\}$.
Each of the ``lifted'' graphical models representations can be viewed as derived from this generic par-factor graph by specifying the language used to express the $\phi_i$-s, the $\anyset{A}_i$-s, and the $\anyset{C}_i$-s.
The probability distribution defined by a par-factor graph is given by the following expression:
\begin{align}
P(\srv{X}  = \srv{x}) &= \anyset{F}(\srv{x}) \notag\\ 
           &= \frac{1}{Z} \prod_{\Phi_i \in \anyset{F}}\Phi_i(\srv{x}) \notag\\
           &= \frac{1}{Z} \prod_{\Phi_i \in \anyset{F}} \prod_{g \in \gnding{\anyset{C}_i}{\Phi_i(\anyset{A}_i)}} g(\srv{x}_g) \label{eqn:parfactor}
\end{align}
The last line considers each possible par-factor in the graph and multiplies together the values of all of its instantiations $g \in \gnding{\anyset{C}_i}{\Phi_i(\anyset{A}_i)}$ on \newstuff{$\srv{x}_g$, the subset of the given assignment \srv{x} needed to evaluate $g$. }In this expression, $g(\srv{x})$ uses just those from the set $\srv{x}$ that are relevant to its computation, i.e. that are the corresponding instantiations to the parameterized RVs in $\anyset{A}_i$. If we compare equations~\ref{eqn:factor} and~\ref{eqn:parfactor}, we see that, indeed, by instantiating a par-factor graph, we obtain a factor graph. However, here, all the factors that are instantiations of the same par-factor share common structure and parameters; thus, par-factor graphs allow for better generalization.

In the remainder of this section, we flesh out this description by considering several popular SRL representations, discussing how they can be viewed as specific instantiations of par-factor graphs, and grouping them according to the type of graphical model they define, i.e. directed, undirected, or hybrid. This is not meant to be an exhaustive list; rather, our goal is to highlight some of the different flavors of representations.
\subsection{Undirected SRL Representations}
This subsection describes SRL representations that can be viewed as ``lifted'' undirected graphical models.
\paragraph{Relational Markov Networks.} As their name suggests, relational Markov networks (RMNs) \cite{taskar:uai02} define Markov networks through a relational representation. RMNs use an object-oriented language and SQL (see Section~\ref{sect:languageterminology}) to specify par-factors. In particular, each par-factor $\Phi = (\anyset{A}, \phi, \anyset{C})$ is defined using an SQL {\tt select} statement in which the {\tt select...from} part establishes the par-RVs in \anyset{A}, and the {\tt where} part establishes the constraints \anyset{C} over instantiations. \newstuff{The RVs resulting from instantiating the par-RVs are multinomial, i.e., they can take on one of multiple discrete values.}  Each tuple returned by the {\tt select} statement constitutes an instantiation of the par-factor; thus, the variables that appear in a returned tuple form a clique in the Markov network defined by the RMN. All of these cliques that are instantiations of the same par-factor share a potential function $\phi$, which takes a log-linear form. In particular, let \srv{A} be a particular instantiation of the par-RVs in \anyset{A}, i.e., \srv{A} is one of the returned tuples. Then, for specific values \srv{a}, assigned to the variables in \srv{A}, $\phi(\srv{A} = \srv{a}) = \exp(\lambda \cdot f(\srv{a}))$, where, as in Markov networks, $f$ is an arbitrary feature function over \srv{A}. Note, however, that unlike in Markov networks, here the potential function $\phi$---its feature function and parameter---is shared across all instantiations of the par-factor $\Phi.$ This property is common to all of the SRL languages we consider here, and is, in fact, one of their main defining characteristics---that through relational languages, they allow for generalization via parameter tying.
 
To illustrate, we provide an example from collective classification of hyperlinked documents, as presented in \cite{taskar:uai02}. The goal of the following par-factor is to set up a clique between the label assignments of any two hyperlinked documents in order to capture the intuition that documents on the web that link to one another typically have correlated labels.
\begin{center}
\begin{small}
\begin{tabular}{p{9cm}}
\begin{verbatim}
select D1.Category, D2.Category
where Document D1, Document D2, Link L
WHERE L.From = D1.Key and L.To = D2.Key
\end{verbatim}
\end{tabular}
\end{small}
\end{center}
This {\tt select} statement sets up the cliques but does not specify the potential function $\phi$ to be used over these cliques. The  definition of $\phi$ can be used to incorporate further domain knowledge in the model. For example, if we know that most pages tend to link to pages of the same category, we can define $\phi(\mathtt{D1.Category}, \mathtt{D2.Category}) = \exp(\lambda \cdot \mathbf{1}[\mathtt{D1.Category} = \mathtt{D2.Category}])$, where the feature function takes the form of the indicator function $\mathbf{1}[\rv{x}]$ that returns 1 if the proposition \rv{x} is true and 0 otherwise. A positive $\lambda$, encourages hyperlinked pages to be assigned the same category, while a negative $\lambda$ discourages this. \newstuff{Taskar et al. \cite{taskar:icml04} have shown that for associative RMNs, in which the factors favor the same labels for RVs in the same clique, inference and learning is tractable (for binary random variables) and closely approximated for the non-binary case.}

\paragraph{Markov Logic Networks.} Markov logic networks (MLNs) \cite{richardson:mlj06,domingos:book09} also define a Markov network when instantiated. Par-factors in MLNs are specified using first-order logic. Each par-factor $\Phi = (\anyset{A}, \phi, \anyset{C})$ is represented by a first-order logic rule $R_{\Phi}$ with an attached weight $\lambda_{\Phi}$. $\anyset{A}$ consists of all par-RVs that appear in the rule; therefore, in the instantiated Markov network, each instantiation, or grounding, of $R_{\Phi}$, establishes a clique among the RVs that appear in that instantiation. \newstuff{The instantiated RVs are Boolean-valued.} The potential function $\phi$ is implicit in the rule, as we describe next. Let \srv{A} be the set of RVs in a particular instantiation, and $\srv{a}$ be a particular assignment of truth values to \srv{A}; then, $\phi(\srv{A} = \srv{a}) = \exp(\lambda_{\Phi} \cdot R_{\Phi}(\srv{a}))$, where $R_{\Phi}(\srv{a}) = 1$ if it is true on the given truth assignment $\srv{a}$ and $R_{\Phi}(\srv{a}) = 0$ otherwise. In other words, clique potentials in MLNs are represented using log-linear functions in which the first-order logic rule itself acts as a feature function, whereas the weight associated with the rule is the parameter.

As an illustration, we present an example from \cite{richardson:mlj06}, in which the patterns of human interactions and smoking habits are considered. One regularity in this domain is that friends have similar smoking habits, i.e., that if two people are friends, then they will either both be smokers or both be non-smokers. This can be captured with the following first-order logic rule (where $\lambda$ is the weight associated with it):
\[ \lambda:  \mathtt{Friends(X, Y)} \Rightarrow (\mathtt{Smokes(X)} \Leftrightarrow \mathtt{Smokes(Y)}) \]
The par-RVs in the par-factor defined by this rule are $\anyset{A} = \{ \mathtt{Friends(X, Y)}, \mathtt{Smokes(X)}, \mathtt{Smokes(Y)}\}$, and every possible instantiation of these par-RVs establishes a clique in the instantiated Markov network, e.g., if there are only two entities, \lv{a} and \lv{b}, then the instantiated Markov network will contain the cliques
\begin{center}
$\{\lv{Friends(a, a), Smokes(a)}\},$\\
$\{\lv{Friends(a, b), Smokes(a), Smokes(b)}\},$\\
$\{\lv{Friends(b, a), Smokes(a), Smokes(b)}\},$ and \\
$\{\lv{Friends(b, b), Smokes(b)}\}$ \\
\end{center}
among the RVs \lv{Smokes(a), Smokes(b), Smokes(a, b), Smokes(b, a),} and \lv{Smokes(b,b)}. 

So far, we have not discussed how MLNs specify the constraints \anyset{C} of a par-factor. MLNs do not have a special mechanism for describing constraints, but constraints can be implicit in the rule structure. Two ways of doing this are as follows. First, we can allow only instantiations that ground a given variable with a specific entity by replacing this variable with the entity name. For example, if we want to constrain $\lv{X} = \lv{a}$ in the rule above, we simply write it as $\mathtt{Friends(a, Y)} \Rightarrow (\mathtt{Smokes(a)} \Leftrightarrow \mathtt{Smokes(Y)})$. A second and more general way of introducing constraints is via predicates whose values will be known at inference time, as in discriminatively trained models. For example, suppose we know that at inference time we will observe as evidence the truth values of all groundings of the $\mathtt{Friends}$ predicate, and the goal will be to infer people's smoking habits. Then, the rule $\mathtt{Friends(X, Y)} \Rightarrow (\mathtt{Smokes(X)} \Leftrightarrow \mathtt{Smokes(Y)})$ can be seen as setting up a clique between the $\mathtt{Smokes}$ values only of entities that are friends. This is because, if for a particular pair of entities \lv{x} and \lv{y}, $\mathtt{Friends(x, y)}$ is false, then the corresponding instantiation of the rule is trivially satisfied, regardless of assignments to groundings of \lv{Smokes}. Therefore, such instantiations can be ignored when instantiating the MLN. 

A variant of MLNs is the Hybird MLN model \cite{wang:aaai08}, which extends MLNs to allow for real-valued predicates. In Hybrid MLNs, the same formula can contain both binary-valued and real-valued terms. Such formulas are evaluated by interpreting conjunction as a multiplication of values.  
\paragraph{Probabilistic Similarity  Logic.} 
Another ``lifted'' Markov network model is probabilistic similarity logic (PSL) \cite{broecheler:uai10}, which allows for reasoning about similarities. In PSL, par-factors are defined as weighted rules expressed as a mix of first-order logic and object-oriented languages. Thus, similar to MLNs, each par-factor  $\Phi = (\anyset{A}, \phi, \anyset{C})$ consists of a rule $R_{\Phi}$ with attached weight $\lambda_{\Phi}$, whose par-RVs determine the set \anyset{A} and whose structure determines the potential function $\phi$. \newstuff{However, unlike in RMNs and MLNs, RVs resulting from instantiating a par-RV are continuous-valued in the interval $[0,1]$. Thus, instantiating a set of PSL rules results in a continuous-valued Markov network.} Constraints in $\anyset{C}$ can be specified in a similar manner as in MLNs.

Unlike previous models, PSL additionally supports reasoning about similarities, both between entity attributes and between sets of entities. Similarity functions can be any real-valued function whose range is the interval $[0,1]$, and in formulas they can be mixed with regular relational terms. To illustrate, consider an example from \cite{broecheler:uai10} in which the task is to infer document similarities in Wikipedia based on document attributes and user interactions with the document. One potentially useful rule states that two documents are similar if the sets of their editors are similar and their text is similar:
\[\{\mathtt{A.editor}\} \stackrel{s_1}{\approx} \{\mathtt{B.editor}\} \wedge \mathtt{A.text} \stackrel{s_2}{\approx} \mathtt{B.text} \Rightarrow \mathtt{A}\stackrel{s_3}{\approx}\mathtt{B} \] 
Above, $\stackrel{s_i}{\approx}$ represent similarity functions, and a term enclosed in curly braces, as in $\{\mathtt{A.editor}\}$, refers to the set of all entities related to the variable through the relation. The par-RVs in the above rule are $\anyset{A} = \{\quad \{\mathtt{A.editor}\} \stackrel{s_1}{\approx} \{\mathtt{B.editor}\}, \mathtt{A.text} \stackrel{s_2}{\approx} \mathtt{B.text}, \mathtt{A}\stackrel{s_3}{\approx}\mathtt{B}\}$, and, as before, the rule defines a clique among each possible instantiation of these par-RVs. 

The evaluation $R_{\Phi}(\srv{a})$ of a rule $R_{\Phi}$ on an assignment \srv{a} to an instantiation \srv{A} of the par-RVs involves combining Boolean values and similarity values. In PSL, this is done using t-norms/t-conorms, which generalize the first-order logic operations of conjunction/disjunction. While any t-norm/t-conorm pair can be used, in \cite{broecheler:uai10} the Lukasiewicz t-(co)norm was preferred due to its property of being linear in the values combined. Letting $x, y \in [0,1]$ be two Boolean or similarity values, the Lukasiewicz t-(co)norm is defined as follows:
\begin{align*}
x \wedge y &= \max\{0, x+y - 1\} \\
x \vee y &= \min\{x+y, 1\} \\
\neg x &= 1-x
\end{align*}
Thus, each instantiated rule has a value in the interval $[0, 1]$, and one can talk about the distance to satisfaction of a rule instantiation and define it as $d(R_{\Phi}(\srv{a})) = 1 - R_{\Phi}(\srv{a})$. \label{distfromsat}

PSL generalizes the standard formulation of the joint distribution (e.g., Equation~\ref{eqn:parfactor}) by interpreting it as a penalty on the distance from satisfaction of a set of PSL rules on a given assignment of values to the instantiated RVs. 
The probability of observing a given joint assignment of values is given by:
\[ P(\srv{X} = \srv{x}) = \frac{1}{Z} \exp(-d_{\delta}(\srv{x}))\]
Above, $\delta$ is an arbitrary distance function, and $d_{\delta}(\srv{a})$ is the distance from satisfaction function, computed as $\delta(V(\srv{a}), \srv{0})$, where $V(\srv{a})$ is the vector of weighted distances to satisfaction of all rule instantiations. If we pick $\delta$ to be the $L_1$-norm distance, $\delta(x,y) = \|x-y\|_1$, we get back the standard formulation, in which the potential function $\phi$ associated with a clique is given by $\phi(\srv{A} = \srv{a}) = \exp(-\lambda_{\Phi} \cdot d(R_{\Phi}(\srv{a})))$. 
\paragraph{Imperatively Defined Factor Graphs.}
Par-factor graphs can also be defined using an imperative programming language, as is done in FACTORIE, an implementation of imperatively defined factor graphs (IDFs) \cite{mccallum:nips09}. FACTORIE uses {\it Scala}, a strongly-typed functional programming language, and in this way provides the model designer with enormous flexibility. In FACTORIE, variable types are represented as typed objects that can be sub-classed, and potential relations between variables of particular types are represented as instance variables in the corresponding classes. In this way, novel types of variables, such as set variables, can be defined, and standard data structures, such as linked lists, can be used in variable type implementations. 

Each par-factor $\Phi = (\anyset{A}, \phi, \anyset{C})$ is defined as a factor template class, whose arguments determine the set of par-RVs \anyset{A}. The factor template class contains a set of $\mathtt{unroll}$ procedures, one for each par-RV, that return the set of instantiations of this par-factor, corresponding to a given instantiation of one of the arguments. The instantiation constraints \anyset{C} are therefore encoded in the $\mathtt{unroll}$ methods. The potential function $\phi$ is implemented via a $\mathtt{statistics}$ method in the factor template class and thus can have arbitrary form (as long as it returns a strictly positive value). \newstuff{Consider a simplified version of an example from \cite{mccallum:nips09} that defines a factor template for the task of coreference resolution. The goal of the factor template is to evaluate the compatibility of an entity $\mathtt{Mention}$ and the canonical representation of the particular underlying $\mathtt{Entity}$ to which it is assigned.  The $\mathtt{unroll1}$ method produces a factor between a given mention and the entity to which it is assigned. Given an entity, the $\mathtt{unroll2}$ method produces a set of factors, one for each mention associated with this entity. }
 \begin{small}
\begin{verbatim}
val corefTemplate = new Template[Mention, Entity]{
   def unroll1(m:Mention) = Factor(m, m.entity)
   def unroll2(e:Entity) = for (mention <- e.mentions) yield Factor(mention, e)
   def statistics(m:Mention,e:Entity) = Bool(distance(m.string,e.canonical)<0.5)
}
\end{verbatim}
\end{small}
\subsection{Directed SRL Representations}
This subsection describes SRL representations that define Bayesian networks when instantiated. Par-factors in these representations have special form. In particular, for a par-factor $\Phi = (\anyset{A}, \phi, \anyset{C})$, \anyset{A} consists of a child par-RV $\rv{C} \in \anyset{A}$ and a set of parent par-RVs $\srv{Pa} = \anyset{A} \setminus \{\rv{C}\}$. Because the function $\phi$ represents a conditional probability distribution for \rv{C} given \srv{Pa}, the expression in Equation~\ref{eqn:parfactor} is automatically normalized, i.e.,  $Z = 1$. When specifying directed SRL models, care must be taken to ensure that their instantiations in different worlds would result in cycle-free directed graphs. However, as discussed in \cite{jaeger:eacis02}, Section~3.2.1, this problem is undecidable in general, and guarantees exist only for restricted cases. 
\paragraph{Bayesian Logic Programs.} In Bayesian logic programs (BLPs) \cite{kersting:ilp01}, par-RVs are expressed as logical atoms, and the dependency structure of \rv{C} on its parents \srv{Pa} is represented as a definite clause, called a Bayesian clause, in which the head consists of \rv{C}, the body consists of \srv{Pa}, and the implication is replaced by a $|$ to indicate probabilistic dependency. A further distinction between ordinary logical clauses and Bayesian clauses is that in the latter, logical atoms are not restricted to evaluating to just $\mathtt{true}$ or $\mathtt{false}$, as illustrated in the example below. Par-factors are formed by coupling a Bayesian clause with a conditional probability distribution (CPD) over values for \rv{C} given values for \srv{Pa}.

Kersting and De Raedt \cite{kersting:ilp01} give an example from genetics (originally from \cite{friedman:ijcai99}), in which the blood type \lv{bt(X)} of person \lv{X} depends on inheritance of a single gene, one copy of which, \lv{mc(Y)} is inherited from \lv{X}'s mother, \lv{mother(Y, X)}, while the other copy \lv{pc(Z)} is inherited from her father, \lv{father(Z, X)}. In BLPs this dependency is expressed as 
\[\lv{bt(X)} | \lv{mc(X)}, \lv{pc(X)} \]
In this example, \lv{mc(X)} and \lv{pc(X)} can take on values from $\{a, b, 0\}$, whereas \lv{bt(x)} can take on values from $\{a, b, ab, 0\}$. The specification of this par-factor is completed by providing for each possible combination of values assigned to  \lv{mc(X)} and \lv{pc(X)}, the probability distribution over values of \lv{bt(x)}, e.g., as a conditional probability table.

Using BLPs, we next illustrate another aspect common to most directed SRL representations: the use of combining rules. Following the example from \cite{kersting:ilp01}, suppose that in the genetics domain we have the following two rules:
\begin{align*}
\lv{bt(X)} &| \lv{mc(X)} \\
\lv{bt(X)} &| \lv{pc(X)} 
\end{align*}
Each of these rules, comes with a CPD. The first one for  \lv{bt(X)} given \lv{mc(X)}, and the second one for  \lv{bt(X)} given \lv{pc(X)}. However, what we need is a single CPD for predicting \lv{bt(X)} given both of these quantities. Such a CPD is obtained by using combining rules, which are functions of one or more CPDs to a single CPD. For example, one commonly used combining function is the noisy-or.
\paragraph{Relational Bayesian Networks.} Relational Bayesian networks (RBNs) \cite{jaeger:eacis02} also represent par-RVs as logical atoms. A separate par-factor $\Phi_R$ is defined for each unknown relation $R$ in the domain, such that the child par-RV $C$ is an atom of $R$. Par-factors are represented as probabilistic formulas in a syntax that bears a close correspondence to first-order logic. The conditional probability distribution of $C$ given \srv{Pa} is implicit in this probabilistic formula, which has range $[0,1]$ and is evaluated as a function of the values of the variables in \srv{Pa}. In particular, probabilistic formulas in RBNs are recursively defined to consist of (i) constants in $[0,1]$, which in the extreme cases of 1 and 0 correspond to $\mathtt{true}$ and $\mathtt{false}$ respectively; (ii) indicator functions, which take tuples of logical variables as arguments and correspond to relational atoms; (iii) convex combinations of formulas, which correspond to Boolean operations on formulas; and, finally, (iv) combination functions, such as $\mathtt{mean}$, that combine the values of several formulas. 

To illustrate, consider a slight adaptation of an example from \cite{jaeger:eacis02}, where the task is, given the pedigree of an individual $\lv{x}$, to reason about the values of two relations,  $\lv{FA(x)}$ and $\lv{MA(x)}$, which indicate whether $\lv{x}$ has inherited a dominant allele $A$ from its father and mother respectively. The probabilistic formula for $\lv{FA(X)}$, may be:
\[\Phi_{\lv{FA}}(\lv{X}) = \Phi_{\lv{knownFather}}(\lv{X}) \cdot \Phi_{\lv{A-from-father}}(\lv{X}) + (1-\Phi_{\lv{knownFather}}(\lv{X})) \cdot \theta \]
Above, $\Phi_{\lv{knownFather}}(\lv{X})$ is an auxiliary sub-formula that evaluates to $\mathtt{true}$ if the father of \lv{X} is included in the pedigree and to $\mathtt{false}$ otherwise; $\Phi_{\lv{A-from-father}}(\lv{X})$ is an auxiliary sub-formula defined as the mean over the $\lv{FA}$ and $\lv{MA}$ values of \lv{X}'s father: $\Phi_{\lv{A-from-father}}(\lv{X}) = \mathtt{mean}\{\lv{FA(Y),MA(Y)} | \lv{father(Y, X)}\}$; and $\theta$ is a learnable parameter that can take values in the range $[0, 1]$. 

As in some of the undirected models discussed earlier, RBNs do not provide a dedicated mechanism for specifying the constraints \anyset{C}. However, constraints may be specified either by replacing logical variables with actual entity names in formulas, or by including tests on the background information, as is the case with the $\Phi_{\lv{knownFather}}$ sub-formula above. 
\paragraph{Probabilistic Relational Models.}
Probabilistic relational models (PRMs) \cite{koller:aaai98,getoor:bkchapter07} take a relational database perspective and use an object-oriented language, akin to that described in Section~\ref{sect:languageterminology}, to specify the schema of a relational domain. Both entities and relations are represented as classes, each of which comes with a set of descriptive attributes and a set of reference slots through which classes refer to one another. Using an example from \cite{getoor:bkchapter07}, consider a document citation domain that consists of two classes, the $\mathtt{Paper}$ class with attributes $\mathtt{Paper.Topic}$ and $\mathtt{Paper.Words}$, and the $\mathtt{Cites}$ class, which establishes a citation relation between two papers via reference slots $\mathtt{Cites.Cited}$ and $\mathtt{Cites.Citing}$. The par-RVs in PRMs correspond to attributes on objects, possibly going through  chains of reference slots. Each par-factor is defined by specifying the par-RVs corresponding to the child node \rv{C} and the parent nodes \srv{Pa} respectively, and providing a conditional probability distribution for \rv{C} given \srv{Pa}. For example, one possible par-factor in the document citation domain may set \newstuff{$\rv{C} = \mathtt{P.Topic}$ and $\srv{Pa} = \{\mathtt{P.Cited.Topic}, \mathtt{P.Citing.Topic}\}$}. Thus, this par-factor establishes a dependency between the topic of a paper and the topics of papers that it cites or that cite it. As with other directed SRL models, PRMs use aggregation functions in cases when the number of parents obtained from instantiating a par-factor varies. For instance, since papers cite varying number of papers, one would need to aggregate over the topics of papers corresponding to the instantiation of $\mathtt{P.Cited.Topic}$.

In the above example, reasoning is performed only about the attributes of objects, whereas their relations are assumed given. However, in some applications it may be necessary to reason about the presence of a relation between two objects. For example, there may be uncertainty regarding $\mathtt{Cites.Citing}$. PRMs provide extensions for dealing with these cases, for two situations: when the number of links is known, but not the specific objects that are linked (reference uncertainty), as well as when neither the number of links nor the linked objects are known (existence uncertainty). 
\paragraph{BLOG.}
BLOG, short for Bayesian LOGic, is a relational language for specifying generative models \cite{milch:ijcai05}.  Par-RVs in BLOG are represented as first-order logic atoms. The dependence of a par-RV on its parents is expressed by listing the parent par-RVs and specifying the distribution from which the child is drawn, given the parents. For example, Milch et al. ~\cite{milch:ijcai05} present a BLOG model for entity resolution. This model views the set of citations of a given paper as being drawn uniformly at random from the set of known publications. This is captured by the following BLOG statement:
\[ \lv{PubCited(C)} ∼\sim \textrm{Uniform}(\mathtt{\{Publication \textrm{ } P\}}). \]
Similarly, the citation string is viewed as being generated by a string corruption model $\mathtt{CitDistrib}$ as a function of the authors and title of the paper being cited:
\[\lv{CitString(C)} \sim∼ \lv{CitDistrib(TitleString(C),AuthorString(C))}\]
A unique characteristic of BLOG is that it does not assume that the set of entities in a domain is known in advance and instead allows reasoning over variable numbers of entities. This functionality is supported by allowing {\it number} statements, in which the number of entities of a given type is drawn from a given distribution. For example, in the entity resolution task, the number of researchers is not known in advance and is instead drawn from a user-defined distribution. Standard distributions such as the Poisson can also be used.

\subsection{Directed versus Undirected Models}
\label{sect:dirvsundir}
Most SRL representations discussed in this survey define either a directed or an undirected graphical model when instantiated. These representations have relative advantages and disadvantages, analogous to those of directed and undirected graphical models \cite{koller:book09}. In terms of representation, directed models are appropriate when one needs to express a causal dependence. On the other hand, undirected models are better suited to domains containing cyclic dependencies. While causal structure might be easier to elicit from experts, when model structure is learned from data, special care needs to be taken with directed models to ensure acyclicity. Structure revisions in directed models can be evaluated much faster for typical scoring functions, which are decomposable, because parameters change locally, only in the places where the dependency structure has changed. In other words, when the set of parents of a node $C$ change, the only parameters that typically need to be adjusted are those of the conditional distribution of $C$ given its new set of parents. This contrasts with undirected models in which scoring the revision of a single par-factor requires adjusting the parameters of the entire model. Issues pertaining to structure learning are further discussed in Section~\ref{sect:structure}. Undirected models have a straightforward mechanism for combining the par-factors shared by a single par-RV, simply by multiplying them, whereas directed models require the use of separately defined \newstuff{combining functions, such as noisy-or, or aggregation functions, such as count, mode,  max, and average}. On the other hand, the use of combining functions in directed models allows for multiple independent causes of a given par-RV to be learned separately and then combined at prediction time \cite{heckerman:uai94}. On the other hand, this kind of causal independence cannot be exploited in undirected models. Finally, because factors in directed models represent conditional probability distributions, they are automatically normalized. In contrast, in undirected models one needs to find efficient ways of computing, or estimating, the normalization constant $Z$ (in Equation~\ref{eqn:parfactor}). 
\subsubsection{Hybrid SRL Representations}
Hybrid SRL representations combine the positive aspects of directed and undirected models. One such model is relational dependency networks (RDNs) \cite{neville:jmlr07}, which can be viewed as a lifted dependency network model.
Dependency networks \cite{heckerman:jmlr00} are similar to Bayesian networks in that, for each variable $X$, they contain a factor $f_X$ that represents the conditional probability distribution of $X$ given its parents, or immediate neighbors, $\srv{Pa}_X$. Unlike Bayesian networks, however, dependency networks can contain cycles and do not necessarily represent a coherent joint probability distribution. Marginals are recovered via sampling, e.g., Gibbs sampling (see Section~\ref{sect:inference}). In this respect, dependency networks are similar to Markov networks, i.e., the set of parents $\srv{Pa}_X$ of a variable $X$ render it independent of all other variables in the network. RDNs \cite{neville:jmlr07} lift dependency networks to relational domains. Par-factors in RDNs are similar to those in PRMs and are represented as conditional probability distributions over values for a child par-RV $C$ and the set of its parents \srv{Pa}. Analogous to dependency networks, however, cycles are allowed and thus, as in dependency networks, RDNs do not always represent a consistent joint probability distribution. 

There has also been an effort to unify directed and undirected models by providing an algorithm that converts a given directed model to an equivalent MLN \cite{natarajan:ecml10}. In this way, one can model multiple causes of the same variable independently while taking advantage of the variety of inference algorithms that have been implemented for MLNs. Bridging directed and undirected models is important also as a step towards representations that combine both directed and undirected sub-components.
\section{Inference}
\label{sect:inference}
An SRL model can be used to draw two types of inferences, which are analogous to those supported by graphical models.
 In the first type, the goal is to compute the marginal probabilities of variables; we will refer to this as {\it computing marginals.} In the second type, MAP (maximum a posteriori) inference, the task is to find the most likely joint assignment of values to the unknown variables, also known as the MAP state.

This section is structured as follows. We start with an overview of algorithms that directly port inference techniques developed in the graphical models literature by instantiating the given SRL model and operating over the resulting factor graph.  Then, we survey ``lifted'' inference approaches that exploit the symmetries present in an SRL model. Stronger emphasis will be placed on approaches developed specifically for SRL representations.
\subsection{Inference on the instantiated factor graph}
\label{sect:unrolledInference}
One advantage of the SRL representations discussed in this survey is that, since they are instantiated to factor graphs, inference algorithms developed in the graphical models literature can be directly ported here. \newstuff{One of the earliest techniques used to efficiently instantiate a given SRL model in a given domain is knowledge-based model construction (KBMC) \cite{wellman:ker92}, which dynamically instantiates a knowledge base only to the extent necessary to answer a particular query of interest. KBMC has been adapted to instantiate both directed, e.g., \cite{koller:uai97,pfeffer:uai99,getoor:jmlr02}, and undirected models, e.g., \cite{richardson:mlj06}. Application of KBMC in these and other frameworks exploit the conditional independency properties implied by the graph structure of the instantiated model; in particular, the fact that in answering a query about a set of random variables \srv{X}, one only needs to reason about variables that are not rendered conditionally independent of \srv{X} given the values of observed variables.  }

Next, we briefly review some of the most commonly used inference techniques. For more detail, we refer the reader to \cite{koller:book09}.

\paragraph{Variable Elimination} One of the earliest and simplest algorithms that can be used to perform exact inference over the grounded factor graph is variable elimination VE \cite{poole:jair03}. Suppose we would like to compute the marginal probability of a particular instantiation \rv{X} of a particular par-RV \rv{X} (the instantiation of a par-RV is an ordinary RV). To do that, we need to sum out all the other variables (i.e. all other instantiations of par-RVs), which we will call \srv{Y}.  VE proceeds in iterations summing out variables from the set \srv{Y} one by one. An ordering over the variables in \srv{Y} is established, and in each iteration the next $\rv{Y} \in \srv{Y}$ is selected, and the set of factors is split into two groups---the ones that contain \rv{Y} and the ones that do not. All factors containing \rv{Y} are multiplied together and \rv{Y} is summed out, thus effectively removing \rv{Y} from \srv{Y}. The efficiency of the algorithm is affected by the ordering over \srv{Y} that was used; heuristics for selecting better orderings are available. In the end the result is normalized. 

This algorithm can be adapted to find the MAP state by maximizing over variables rather than summing over them. In particular, suppose we would like to find the most likely joint assignment to a set of variables \srv{X}. As before, we impose an ordering over \srv{X} and proceed in iterations, this time, however, eliminating each variable $\rv{X} \in \srv{X}$  by {\it maximizing} the product of all factors that contain \rv{X} and remembering which value of \rv{X} gave the maximum value. 
\paragraph{Belief Propagation} Another algorithm that can be used to compute marginals over a factor graph is belief propagation (BP) \cite{pearl:book88}, also known as the sum-product algorithm \cite{kschischang:infotheory01} because it consists of a series of summations and products. BP computes the marginals exactly on graphs that contain no cycles. A special case that frequently arises in practice is that of chain graphs, in which BP is known as the forward-backward algorithm \cite{rabiner:ieee89}. BP's operation is based on a series of ``messages'' that are sent from variable nodes to factor nodes and vise versa. For a complete derivation of these messages, we refer the reader to \cite{kschischang:infotheory01}; here we provide only the result, using the notation in that paper. Messages sent from a variable node $X$ to a factor node $f$ are denoted \mess{X}{f}{X}, and messages sent from a factor node $f$ to a variable node $X$ are denoted \mess{f}{X}{X} (note that both are functions of $X$). The messages sent are defined as shown below. In the following expressions, $n(X)$ denotes the set of neighbors of node $X$, i.e. these are the factors in which it participates; analogously, $n(f)$ is the set of variables used to calculate a factor; and $\sim\{X\}$ denotes the set of all variables except $X$:
\begin{equation}
\mess{X}{f}{X} = \prod_{h \in n(X)\setminus\{f\}}\mess{h}{X}{X}
\label{eqn:BPxtof}
\end{equation} 
\begin{equation}
\mess{f}{X}{X} = \sum_{\sim\{X\}}\left(f(n(f)) \prod_{Y \in n(f) \setminus \{X\}} \mess{Y}{f}{Y}\right)
\label{eqn:BPftox}
\end{equation}
In other words, before sending a message to a neighbor, a variable waits to receive messages from all of its other neighbors, and then simply sends the product of these messages. If the variable is a leaf node, thus having a single neighbor, it sends it a trivial message of 1. Similarly, before a factor sends a message to a variable, it waits to receive messages from all other variables and then multiplies these messages together with itself and sums out all other variables except for the one to which the message is sent. If the factor is a leaf node, it simply sends itself. In the end, the marginal at a variable $X$ is calculated as the product of all incoming messages from the neighboring factors.  
BP can also be used on loopy graphs where it is run for a sequence of iterations. Although in such cases it is not guaranteed to output correct results, in practice it frequently converges and, when this happens, the values obtained are typically correct \cite{murphy:uai99,yedidia:ijcai01}. 

As in variable elimination, BP can be easily adapted to compute the MAP state by replacing the summation operator in Equation~\ref{eqn:BPftox} with a maximization operator. This is called the max-product algorithm, or, if the underlying graph is a chain, the Viterbi algorithm.
\paragraph{Sampling}
Exact inference is intractable in general. An alternative approach is to perform approximate inference, based on sampling. One of the most popular techniques is Gibbs sampling, a Markov chain Monte Carlo (MCMC) sampling algorithm.  At the onset, Gibbs sampling initializes the unknown variables. This can be done randomly,  but faster convergence may be obtained with more carefully picked values, i.e. ones that result in a MAP state. Sampling then proceeds in iterations, where in each iteration a new value for one of the unknown variables is sampled, given the current assignments to the remaining variables. Under general conditions, Gibbs sampling converges to the target distribution \cite{tierney:as94}. 

One of these conditions that may often be broken in practice is ergodicity, or the requirement that each state (i.e. particular configuration of assignments to the variables) can be aperiodically visited from each other state. Ergodicity is violated when the domain contains deterministic or near-deterministic dependencies, in which case sampling becomes stuck in one region and converges to an incorrect result.  One way of avoiding this problem is to jointly sample new values for blocks, or groups, of variables with closely coordinated assignments. Another way of overcoming the challenge of deterministic or near-deterministic dependencies is to perform slice sampling \cite{damien:jrss99}. Intuitively, in slice sampling, auxiliary variables are used to identify ``slices'' that ``cut'' across the modes of the distribution. By sampling uniformly from a slice, this technique allows sampling to jump across modes, thus preventing it from getting stuck in a single region. Slice sampling was used to derive the MC-SAT algorithm \cite{poon:aaai06}, which performs slice sampling in MLNs. MC-SAT identifies the slice from which to sample as the set of all possible variable assignments that satisfy an appropriately selected set of grounded clauses (which in MLNs define the factors), where the probability of selecting a grounded clause is larger for clauses with larger weights. MC-SAT samples (near-) uniformly from this slice using the SampleSAT algorithm \cite{wei:aaai04}.

An orthogonal concern is the efficiency of sampling. One approach to speeding up sampling is to use memoization \cite{pfeffer:aaai07}, in which values of past samples are stored and reused, instead of generating a new sample. If care is taken to keep reuses independent of one another, the accuracy of sampling can be improved by allowing the sampler to draw a larger number of samples in the allotted time. 
\paragraph{Weighted Satisfiability}
MAP inference with MLNs is equivalent to finding a joint assignment to the par-RV instantiations such that the weight of satisfied formula instantiations is maximized. In other words, performing MAP inference in MLNs is equivalent to solving a weighted satisfiability problem, as discussed by Richardson and Domingos \cite{richardson:mlj06} who used the MaxWalkSat algorithm \cite{kautz:dimacs97}.
\paragraph{(Integer) Linear Programming}
MAP inference can be performed by also solving an integer linear program that is constructed from the given factor graph. In the most general construction, e.g., \cite{koller:book09,taskar:thesis04}, a Boolean-valued variable $v_f^{\mathbf{j}}$ is introduced into the program for each factor $f$ and each possible assignment of values $\mathbf{j}$ to the RVs $\srv{X}_f$ involved in the computation of $f$. Constraints are added to enforce the conditions that (1) for each factor $f$ only one of the $v_f^{\mathbf{j}}$ is set to 1 at any given time and (2) the values of $v_{f_1}^{\mathbf{j}}$ and $v_{f_2}^{\mathbf{j'}}$ where $f_1$ and $f_2$ share variables are consistent. The MAP state is found by maximizing the objective
$\log \prod_{f} f(\mathbf{x}_f) = \sum_{f, \mathbf{j}} v_f^{\mathbf{j}} f(\mathbf{x}_f = \mathbf{j})$ under these constraints.

Specializations of this general procedure of generating constraints for a factor graph exist for the cases of factor graphs obtained by instantiating MLNs and PSL. A specialized procedure for casting MAP inference in MLNs as an integer linear program is provided in \cite{riedel:uai08}.  The resulting linear program contains a variable $y_X$ for each par-RV instantiation \rv{X} whose value is unknown, as well as a variable $\lambda_f$ for each instantiation of each formula (corresponding to each factor in the instantiated factor graph). The formula instantiations are simplified by replacing par-RVs whose values are known with their observed values (e.g., $\mathtt{true}$ or $\mathtt{false}$) and replacing all other par-RV instantiations with their corresponding variables $y$. The correspondence between a formula instantiation $f$ and its corresponding variable $\lambda_f$ is established by requiring logical equivalence between $f$ and $\lambda_f$, effectively rewriting each $f$ as $\lambda_f \Leftrightarrow f$. These rewritten formulas are then converted to conjunctive normal form, and each disjunction transformed into a linear constraint. As an example of how this latter step is carried out, consider the disjunction $\neg X \vee Y$. Its corresponding linear constraint is $-1.0 X + 1.0 Y \geq 0$.

In PSL, MAP inference is cast as a second-order cone program. This is done as follows\cite{broecheler:uai10}. As with MLNs, the program contains a variable $v_X$ for each unknown par-RV instantiation $X$ and a rule variable $v_R$ for each PSL rule instantiation $R$. All par-RV instantiations are replaced with their corresponding variables, and the correspondence between rule variables and their corresponding instantiated rules is established by including the constraint $v_R \geq d(R(\srv{a}))$, where $d(R(\srv{a}))$ represents the distance to satisfaction of rule instantiation $R$, as defined in the description of PSL on page~\pageref{distfromsat}. Additional (hard) constraints can also be included. Because par-RV instantiations in PSL have continuous, rather than Boolean, values,  the variables in the optimization are not constrained to be integers, as is the case in MLNs. As a result, under appropriate choices for t-(co)norms and similarity functions, the second-order cone program can be solved efficiently in polynomial time, as opposed to the integer linear programming case, which is known to be NP-hard \cite{schrijver:ilpbook98}.

\subsection{Improving Inference Efficiency}
\paragraph{Cutting Plane Inference} As we described them above, the procedures for casting MAP inference in MLNs and PSL as an optimization problem are naive in the sense that they first fully instantiate the given formulas or rules and then convert them to constraints. In reality, the efficiency of both procedures is significantly improved by making use of a {\it default} value --- $\mathtt{false}$ in the case of MLNs, and 0.0 in the case of PSL --- and realizing that most rule instantiations, and thus their corresponding constraints, will be satisfied by setting par-RV instantiations to their default value. In other words, the only constraints that need to be included in the optimization problem are those corresponding to rule instantiations that are not yet fully satisfied by the current assignment of values to the unknown par-RV instantiations. This realization leads to an iterative procedure  \cite{riedel:uai08,broecheler:uai10}, whereby a series of optimization problems are solved, each subsequent problem including only those additional constraints that are not yet fully satisfied by the assignment of values to the par-RV instantiations. Riedel \cite{riedel:uai08} relates this procedure to cutting plane algorithms developed in the operations research community.  In cutting plane optimization, rather than solving the original problem, one first optimizes a smaller problem that contains only a subset of the original constraints. The algorithm then proceeds in iterations, each time adding to the active constraints those from the original problem that are not satisfied by the current solution. This process continues until a solution that satisfies all constraints is found. In the worst case, this may require considering all constraints. However, in many cases, it will be possible to find a solution by only considering a small subset of the constraints. 

\paragraph{Lazy Inference} Lazy inference is a related meta-inference technique that is based on the fact that relational domains are typically sparse, i.e., \newstuff{that very few of all possible relations are actually true.} Lazy inference was originally developed to improve the memory efficiency of MAP inference with MaxWalkSat for MLNs, resulting in the LazySAT algorithm \cite{singla:aaai06}. LazySat maintains sets of {\it active} RVs and {\it active} formula instantiations. Only the active RVs and formulas are explicitly maintained in memory, thus dramatically decreasing the memory requirements of inference. Initially, all RVs are set to $\mathtt{false}$, and the set of active RVs consists of all RVs participating in formula instantiations that are not satisfied by the initial assignment of $\mathtt{false}$ values. A formula instantiation is activated if it can be made unsatisfied by flipping the value of zero or more active RVs. Thus the initial set of active formula instantiations consists of those activated by the initially active RVs. The algorithm then carries on with the iterations of MaxWalkSat, activating RVs when their value gets flipped and then activating the relevant rule instantiations. LazySat was later generalized to other inference techniques, such as sampling \cite{poon:aaai08}.

\paragraph{Faster Instantiation}
The efficiency of inference with the techniques described in Section~\ref{sect:unrolledInference} is affected by how quickly the par-factor graph is grounded to a factor graph. For the case of MLNs, Shavlik and Natarajan \cite{shavlik:ijcai09} introduced FROG,  an algorithm that preprocesses a given MLN to improve the efficiency of instantiation. The basic idea is that if an evidence literal in a grounded clause is satisfied, then the clause is already $\mathtt{true}$ regardless of the values of the remaining literals and can be excluded from consideration; thus, when grounding a clause, one only needs to consider variable substitutions that lead to evidence literals not being satisfied, which, in many cases results in a significantly lower number of instantiations. FROG employs a set of heuristics to identify groups of variable substitutions that can be safely ignored. 
\subsection{Lifted Inference}
\label{sect:liftedInference}
By performing inference on the instantiated factor graph, one can take advantage of the many available inference techniques that have been extensively studied in the graphical models literature. However, particularly on larger problems, such an approach can be prohibitive in terms of both memory requirements and running time. To address this issue, a variety of ``lifted'' inference techniques have been developed. Lifted inference exploits the observation that factor graphs obtained by grounding a set of par-factors exhibit a large degree of symmetry, which would lead to repeating the same set of computations multiple times. By organizing these computations in a way that exploits the symmetries and avoids repetition, lifted inference techniques can lead to large efficiency gains.  \newstuff{The earliest techniques are based on recognizing identical structure that would result in repeated identical computations, and performing the computation only the first time, caching the results and subsequently reusing them  \cite{koller:uai97,pfeffer:uai99}. Next we describe several lifted inference approaches, organizing them according to the underlying inference algorithm they use.}
\paragraph{Lifted Variable Elimination}
\newstuff{First-order variable elimination (FOVE) was introduced by  \cite{poole:ijcai03} and later significantly extended in a series of works \cite{desalvobraz:ijcai05,desalvobraz:aaai06,milch:aaai08}}. As in ordinary VE, the goal of FOVE is to obtain the marginal distribution over a set of variables \srv{Q} by summing out the values of the remaining ones. However, unlike VE, FOVE sums over entire sets of variables, such as the possibly constrained set of all groundings of a  par-RV. Thus, the crux of FOVE is to define operations that eliminate, or sum out, entire sets of groundings of par-RVs in such a way that the result is the same as the one that would be obtained by summing out each RV individually. Next we provide a brief discussion of the several elimination operations that have been defined and summarize the conditions under which they apply. For a more detailed treatment, we refer the reader to the above papers; a more unified treatment is presented in \cite{desalvobraz:bkchapter07} and an excellent basic introduction with examples is presented in \cite{kisynski:uai09}. All of the elimination operations assume the following two conditions on the model and use auxiliary operations to achieve them. First, the par-factors in the model need to be {\it shattered} \cite{desalvobraz:ijcai05}, which means that for any two par-factors in the model, the sets of groundings of their par-RVs under the constraints are either identical or completely disjoint. Intuitively, this is necessary in order to ensure that the same reasoning steps will apply to all of the factors resulting from grounding a given par-factor. A model is shattered using  the {\it splitting} operation \cite{poole:ijcai03}. Second, there should be just one par-factor in the model containing the par-RV that is being eliminated. This is achieved using the fusion operation \cite{desalvobraz:ijcai05} which essentially multiplies together all par-factors that depend on the par-RV being eliminated. To facilitate the remainder of this discussion, let \lv{q} be the par-RV we are trying to eliminate, and $\phi$ be the par-factor that depends on \lv{q}.
\subparagraph{Elimination Operators}The first elimination operation is {\it inversion elimination}, introduced by \cite{poole:ijcai03} (the conditions for its correctness were completely specified in \cite{desalvobraz:ijcai05}). Inversion elimination applies only when a one-to-one correspondence can be established between the groundings of \lv{q} and those of $\phi$. This condition is violated when the logical variables that appear in \lv{q} are different from the logical variables in $\phi$. For example, suppose that \lv{q} depends on the logical variable \lv{X} and $\phi$ depends on the par-RVs $\{\lv{p(X, Y)}, \lv{q(X)}\}$. Inversion elimination would not work in this case because \lv{q} does not depend on the logical variable \lv{Y}. In a nutshell, inversion elimination takes a sum of products, where the sum is over all groundings of \lv{q} and the products are over all possible substitutions to the variables in $\phi$ and simplifies it to a product of sums, where each sum now depends on the number of possible truth assignments to \lv{q} (e.g., {\tt true} or {\tt false}), which is in general much smaller than the total number of its groundings. 

Another elimination operation is {\it counting elimination} \cite{desalvobraz:ijcai05}, which is based on the insight that frequently the factors resulting from grounding $\phi$ form a few large groups with identical members. These groups can be easily identified by considering the possible truth assignments to the groundings of $\phi$'s arguments. For each possible truth assignment to the grounded arguments, counting elimination counts the number of groundings that would have that truth assignment. Then only one grounding from each group needs to be evaluated and the result exponentiated to the total number of groundings in that group. To be able to count the number of groundings resulting in a particular truth assignment efficiently, counting elimination requires that the choice of substitutions when grounding one of the par-RVs of $\phi$ does not constrain the choice of substitutions for any of the other ones. Finally, although we have described counting elimination in the context of eliminating the groundings of just one par-RV (\lv{q}), in fact it can be used to eliminate a set of par-RVs.

Finally, \cite{milch:aaai08} introduced elimination {\it with counting formulas} which exploits exchangeability between the parameterized random variables on which a given par-factor depends.  In particular, this property is observed when the par-factor is a function of the number of arguments with a particular value rather than the precise identity of these arguments. The extended algorithm is called C-FOVE.

As we discussed in Section~\ref{sect:dirvsundir}, directed models may require aggregation over a set of values. This may happen, for example, when there is a par-factor in which the parent par-RVs contain variables that do not appear in the child par-RV. In order to aggregate over such variables in a lifted fashion,  \cite{kisynski:ijcai09} introduced {\it aggregation par-factors} and defined a procedure via which an aggregation par-factor is converted to a product of two par-factors, one of which involves a counting formula. In this way, they are able to handle aggregation using C-FOVE. 
\subparagraph{Starting from an Instantiated Graph }FOVE and its extensions apply when an {\it ungrounded} par-factor graph is provided, and their goal during inference is to consider as few grounded cases as possible. An alternative setting is when an {\it instantiated} factor graph is given, and the goal is to recognize the symmetries that are present in it in order to avoid repeated computation. Thus, in the first setting the potential symmetries are implicit in the specification of the ungrounded model and the task is to find out how the presence of evidence breaks these symmetries, whereas in the second scenario only the grounded model is provided, and the task is to recover the symmetries. This latter case naturally occurs when querying a probabilistic database, e.g., \cite{sen:vldbj09}, and was studied by \cite{sen:vldb08}. Their method identifies {\it shared} factors, which compute the same function and is based on the observation that if the inputs of two shared factors have the same values, their outputs will also be the same. Shared factors are discovered by constructing an {\it rv-elim} graph, which simulates the operation of VE without actually computing factor values. The {\it rv-elim} graph provides a convenient way of identifying shared factors whose computations can be carried out once and cached as needed for later. 

All lifted VE algorithms described so far carry out exact computations. Speed of inference can be further improved by performing approximate inference. \cite{sen:uai09} extended their algorithm to this setting by relaxing the conditions for considering two factors to be the shared. One way of doing this is by declaring two factors to be shared if the last $k$ computations as simulated in the {\it rv-elim} graph are the same. Thus, this approach is based on the intuition that the effect of more distant influences is relatively small. Another approximation scheme used by \cite{sen:uai09} is to place factors together in bins if the values they compute are closer than some threshold.
\paragraph{Lifted Belief Propagation}
Lifted BP algorithms \cite{jaimovich:uai07,singla:aaai08,kersting:uai09,desalvobraz:srl09} proceed in two stages. In the first stage, the grounded factor graph $F$ is compressed into a ``template'' graph $T$, in which super-nodes represent groups of variable or factor nodes that send and receive the same messages during BP. Two super-nodes are connected by a super-edge if any  of their respective members in $F$ are connected by an edge, and the weight of the super-edge equals the number of ordinary edges it represents. Once the template graph $T$ is constructed, BP is run over it with trivial modifications. The message sent from a variable super-node $X$ to a factor super-node $f$ is given by 
\begin{equation}
\mess{X}{f}{X} = \mess{f}{X}{X}^{w(X,f) - 1} \prod_{g \in \{n(X) \setminus f\}} \mess{g}{X}{X} ^ {w(X, g)}
\label{eqn:LBPxtof}
\end{equation}
In the above expression, $w(X, f)$ is the weight of the super-edge between $X$ and $f$, and $n(X)$ is the set of neighbors of $X$. The message sent from a factor super-node $f$ to a variable super-node $X$ is given by
\begin{equation}
\mess{f}{X}{X} = \sum_{\sim\{X\}} \left(f(n(f)) \mess{X}{f}{X}^{w(f, X) -1} \prod_{Y \in n(f) \setminus\{ X\}} \mess{Y}{f}{Y}^{w(Y, f)} \right)
\label{eqn:LBPftox}
\end{equation}
At this point, we invite the reader to compare equations~\ref{eqn:LBPxtof} and~\ref{eqn:LBPftox} to their counterparts in the standard case, equations~\ref{eqn:BPxtof} and~\ref{eqn:BPftox}. We note that the lifted case is almost identical to the standard one, except for the super-edge weight exponents.

Next, we describe how the template factor graph is constructed. The first algorithm was given by Jaimovich \etal\ \cite{jaimovich:uai07}. This algorithm targets the scenario when no evidence is provided and is based on the insight that in this case, factor nodes and variable nodes can be grouped into types such that two factor/variable nodes are of the same type if they are groundings of the same par-factor/parameterized variable. The lack of evidence ensures that the grounded factor graph is completely symmetrical and any two nodes of the same type have {\it identical} local neighborhoods, i.e. they have the same numbers of neighbors of each type. As a result, using induction on the iterations of loopy BP, it can be seen that all nodes of the same type send and receive identical messages. As pointed out by the authors \cite{jaimovich:uai07}, the main limitation of this algorithm is that it requires that no evidence be provided, and so it is mostly useful during learning when the data likelihood in the absence of evidence is computed.

Singla and Domingos \cite{singla:aaai08} built upon the  algorithm of Jaimovich \etal\ \cite{jaimovich:uai07} and introduced lifted BP for the general case when evidence is provided. In the absence of evidence, their algorithm reduces to that of Jaimovich \etal\
In this case, the construction of the template graph is a bit more complex and proceeds in stages that simulate BP to determine how the propagation of the evidence affects the types of messages that get sent. Initially, there are three variable super-nodes containing the true, false, and unknown variables respectively. In subsequent iterations, super-nodes are continually refined as follows. First, factor super-nodes are further separated into types such that the factor nodes of each type are functions of the same set of variable super-nodes. Then the variable super-nodes are refined such that variable nodes have the same types if they participate in the same numbers of factor super-nodes of each type. This process is guaranteed to converge at which point the minimal (i.e. least granular) template factor graph is obtained. 

Kersting \etal \cite{kersting:uai09} provide a generalized and simplified description of \cite{singla:aaai08}'s algorithm, casting it in terms of general factor graphs, rather than factor graphs defined by probabilistic logical languages, as was done by Singla and Domingos. Finally, lifted BP has been extended for the any-time case \cite{desalvobraz:srl09}, which combines the approach of \cite{singla:aaai08} with that of \cite{mooij:nips08}. 
\paragraph{Clustering} 
An alternative approach is to cluster the RVs in the instantiated factor graph by the similarity of their neighborhoods, and then compute the marginal probability of only one representative of each cluster, assigning the result to all members of a cluster. This approach was taken with the BAM algorithm \cite{mihalkova:ilp09}, where the neighborhood of each RV $X$ was restricted by setting the values of RVs at a given distance from $X$ to their MAP values, thus cutting off the influence of neighbors that are further away. RVs whose neighborhoods to the given depth are identical are then clustered.
\section{Learning} 
Analogous to learning of graphical models, learning of par-factor graphs can be decomposed into structure learning and parameter learning. Structure learning entails discovering the dependency structure of the model, i.e., what par-RVs should participate together in par-factors. On the other hand, parameter learning involves finding an appropriate parameterization of these par-factors. Parameter learning typically has to be performed multiple times during structure learning in order to score candidate structures. In some applications, the designer hard-codes the structure of the par-factors as background knowledge, and training consists only of parameter learning.
\subsection{Parameter Learning}
\label{sect:weight}
Algorithms for parameter learning of graphical models can be extended in a straightforward way for parameter learning of ``lifted'' graphical models. This extension is based on the realization that an instantiated par-factor graph is simply a factor graph in which subsets of the factors, namely the ones that are instantiations of the same par-factor, have identical parameters. The standard terminology for such par-factors that share their parameters is to say that their parameters are {\it tied}. Thus, in its most basic form, parameter learning in par-factor graphs can be reduced to parameter learning in factor graphs by forcing factors that are instantiations of the same par-factors to have their parameters tied. 

While a complete treatment of parameter learning in graphical models is beyond the scope of this survey, we next provide a brief overview of basic approaches and discuss how they can be easily extended to allow for learning with tied parameters. For a detailed discussion, we refer the reader to \cite{koller:book09}. 

We consider the case of fully observed training data. Maximum likelihood parameter estimation (MLE) aims at finding values for the parameters such that the probability of observing the training data \anyset{D} is maximized, i.e., we are interested in finding values $\srv{\theta}^*$, such that $\srv{\theta}^* =  \arg \max_{\srv{\theta}} P(\anyset{D} | \anyset{F}_{\srv{\theta}})$. For now, let $\anyset{F}_{\srv{\theta}}$ be a factor graph parameterized by $\srv{\theta}$. We now find it helpful to consider directed and undirected models separately. In the directed case, e.g., Bayesian networks, parameter learning consists of learning a conditional probability distribution (CPD) function for each node given its parents. Thus, in the simplest scenario, $\srv{\theta}$ consists of a set of conditional probability tables , one for each node. The maximum likelihood estimate for the entry of a node \rv{A} taking on a value \rv{a}, given that it's parents \srv{B} have values \srv{b}, is found simply by calculating the proportion of time that configuration of values is observed in \anyset{D}:
\begin{equation}
P_{\anyset{D}}^{MLE}(A = a | \srv{B} = \srv{b}) = \frac{\mathtt{count}_{\anyset{D}}(A = a, \srv{B} = \srv{b})}{\sum_{a'}\mathtt{count}_{\anyset{D}}(A = a', \srv{B} = \srv{b})}
\label{eqn:BNs}
\end{equation}
In undirected models, the situation is slightly more complicated because the MLE parameters cannot be calculated in closed form, and one needs to use gradient descent or some other optimization procedure. Supposing that, as introduced in Section~\ref{sect:graphicalmodels}, our representation is a log-linear model with one parameter per factor, then the gradient with respect to a parameter $\theta_i$ of a potential function $\phi_i$ is given by:
\begin{equation}
\frac{\partial}{\partial \theta_i} = \phi_i(\mathcal{D}) - \mathbb{E}_{\theta_i}(\phi_i)
\label{eqn:MNs}
\end{equation}
Above, $\mathbb{E}_{\theta_i}(\phi_i)$ is the expected value of $\phi_i$ according to the current estimate for $\theta_i$.

We next describe how Equations~\ref{eqn:BNs} and~\ref{eqn:MNs} are extended to work with tied parameters. In a nutshell, this is done by computing statistics over all factors that share a set of parameters. In directed models, factors with tied parameters share their CPDs. Thus, in this case, in Equation~\ref{eqn:BNs} counts are computed not just for a single node but for all nodes that share the CPD. Let \srv{A} be that set of nodes, and let $\srv{B}_A$ be the set of parents of node \rv{A}. Then for all $A \in \srv{A}$, Equation~\ref{eqn:BNs} becomes:
\begin{equation}
P_{\anyset{D}}^{MLE}(A = a | \srv{B} = \srv{b})  = \frac{\sum_{A \in \srv{A}} \mathtt{count}_{\anyset{D}}(A = a, \srv{B}_A = \srv{b})}{ \sum_{A \in \srv{A}}\sum_{a'}\mathtt{count}_{\anyset{D}}(A = a', \srv{B}_A = \srv{b})}
\label{eqn:rBNs}
\end{equation}
Analogously, in the undirected case, Equation~\ref{eqn:MNs} is modified to sum over all potentials in the set $\Phi_{\theta_i}$ that share a parameter $\theta_i$:
\begin{equation}
\frac{\partial}{\partial \theta_i} = \sum_{\phi \in \Phi_{\theta_i}}\phi(\mathcal{D}) - \mathbb{E}_{\theta_i}\left(\sum_{\phi \in \Phi_{\theta_i}}\phi\right)
\label{eqn:rMNs}
\end{equation}

\newstuff{One issue that arises when learning the parameters of an SRL model as described above is computing the sufficient statistics, e.g., the counts in Equation~\ref{eqn:rBNs} and the sums in Equation~\ref{eqn:rMNs}. Models that are based on a database representation can take advantage of database operations to compute sufficient statistics efficiently. For example, in PRMs, the computation of sufficient statistics is cast as the construction of an appropriate view of the data and then running simple database queries on it \cite{getoor:thesis02}. Caching is used to achieve further speed-ups. }

\newstuff{While the above discussion focused on one particular learning criterion, that of maximum likelihood estimation, in practice other criteria exist. For example, rather than optimizing the data likelihood, one can dramatically improve efficiency by instead optimizing the pseudo-likelihood \cite{besag:statistician75}. The pseudo-likelihood is computed by multiplying the probability of each variable, conditioned on the values of its Markov blanket observed in the data. As another example, to reduce overfitting, one may use Bayesian learning and impose a prior over the learned parameters \cite{heckerman:bkchapter99,koller:book09}. }

\newstuff{Parameter learning in PRMs, both with respect to a maximum likelihood criterion and a Bayesian criterion, is discussed in \cite{getoor:thesis02}. }
Lowd and Domingos present a comparison of several parameter learning methods for MLNs \cite{lowd:pkdd07}. A max-margin parameter learning approach is presented in \cite{huynh:ecml09} and later extended to train parameters in an online fashion \cite{huynh:sdm11}.
\subsection{Structure Learning}
\label{sect:structure}
\begin{algorithm}[t]
\caption{Generic Structure Learning Algorithm}
\label{alg:genericSL}
\begin{algorithmic}[1]
\INPUT 
\STATEA \anyset{H}: Hypothesis space, possibly encoding language bias\\
\STATEA $\mathbb{A}$: Search algorithm\\
\STATEA $\rho$: Refinement operator\\
\STATEA $\mathbb{S}$: Scoring function
\OUTPUT 
\STATEA Set of par-factors
\PROCEDURE
\STATE $\anyset{S}_0 \gets \mathtt{generateInitialCandidates()}$
\WHILE { $\mathtt{moreTime} \&\& \mathtt{observeImprovements}$}
    \STATE $\anyset{S}_i \gets \mathtt{generateCandidateRefinements}(\anyset{S}_{i-1}, \rho, \mathbb{A}, \anyset{H})$
    \FOREACH {$s \in \anyset{S}_i$}
       \STATE Add $\mathbb{S}(s)$ to $\mathtt{Scores}(\anyset{S}_i)$
    \ENDFOR
    \STATE $\anyset{S}_i \gets \mathtt{Prune}(\anyset{S}_i, \mathtt{Scores}(\anyset{S}_i), \mathbb{A})$ 
\ENDWHILE
\STATE Return $\mathtt{bestModel}(S_{last}, \mathbb{A}, \mathbb{S})$
\end{algorithmic}
\end{algorithm}
The goal of structure learning is to find the skeleton of dependencies and regularities that make up the set of par-factors. Structure learning in SRL builds heavily on corresponding work in graphical models and inductive logic programming (ILP). The basic structure learning procedure can be viewed as a greedy heuristic search through the space of possible structures. A generic structure learning procedure is shown in Algorithm~\ref{alg:genericSL}. This procedure is parameterized by a hypothesis space \anyset{H}, which could potentially encode a language bias; a search algorithm $\mathbb{A}$, e.g., beam search; a refinement operator $\rho$, which specifies how new structures are derived from a given one; and a scoring function $\mathbb{S}$, which assigns a score to a given structure. The algorithm starts by generating an initial set of candidates. This typically consists of trivial par-factors, e.g. ones consisting of single par-RVs. It then proceeds in iterations, in each iteration refining the existing candidates, scoring them, and pruning from the current set of candidates ones that do not appear promising. The details of how these steps are carried out depend on the particular choices for  $\mathbb{A}$, $\rho$, and $\mathbb{S}$. 
The refinement operator $\rho$ is language-specific, but it typically allows for several kinds of simple incremental changes, such as the addition or removal of parents of a par-RV. Common choices for $\mathbb{S}$ are (pseudo) log-likelihood or related measures that can be directly ported from the graphical models literature. Algorithm~\ref{alg:genericSL} is directly analogous to approaches for learning in graphical models, such as \cite{dellapietra:pami97,heckerman:bkchapter99}, as well as to approaches developed in ILP, such as the {\sc foil} algorithm \cite{quinlan:mlj90}. Variants of this algorithm, adapted to the particular SRL representation, have been used by several authors. \newstuff{For example, for the directed case, an instantiation of this algorithm to learning PRMs is described in \cite{getoor:thesis02}. In this case, the $\mathtt{generateCandidateRefinements}$ method checks for acyclicity in the resulting structure and employs classic revision operators, such as adding, deleting, or reversing and edge. In addition to the greedy hill-climbing algorithm that always prefers high-scoring structures over lower-scoring ones, Getoor~\cite{getoor:thesis02} presents a randomized technique with a simulated annealing flavor where at the beginning of learning the structure search procedure takes random steps with some probability $p$ and greedy steps with probability $1-p$. As learning progresses, $p$ is decreased, gradually steering learning away from random choices.
 For the case of undirected models,} Kok and Domingos \cite{kok:icml05} introduced a version of this algorithm for learning of MLN structure. Their algorithm proceeds in iterations, each time searching for the best clause to add to the model. Searching can be performed using one of two possible strategies--beam search or shortest-first search. If beam search is used, then the best $k$ clause candidates are kept at each step of the search. On the other hand, with shortest-first search, the algorithm tries to find the best clauses of length $i$ before it moves on to length $i+1$. Candidate clauses in this algorithm are scored using the weighted pseudo log-likelihood measure, an adaptation of the pseudo log-likelihood that weighs the pseudo likelihood of each grounded atom by 1 over the number of groundings of its predicate to prevent predicates with larger arity from dominating the expression.

Another technique developed in the graphical models community that has been extended to par-factor graphs is that of structure selection through appropriate regularization. In this approach \cite{lee:nips06}, a large number of factors of a Markov network are evaluated at once by training parameters over them and using the $L_1$ norm as a regularizer (as opposed to the typically used $L_2$ norm). Since the $L_1$ norm imposes a strong penalty on smaller parameters, its effect is that it forces more parameters to 0, which are then pruned from the model. Huynh and Mooney \cite{huynh:icml08} extended this technique for structure learning of MLNs by first using Aleph \cite{srinivasan:01}, an off-the-shelf ILP learner, to generate a large set of potential par-factors (in this case, first-order clauses), and then performed $L_1$-regularized parameter learning over this set. 

One of the difficulties of structure learning via greedy search, as performed in Algorithm~\ref{alg:genericSL}, is that the space over possible structures is very large and contains many local maxima and plateaus. Thus, much work has focused on developing approaches that address this challenge. Next, we discuss two groups of approaches. The first group exploits more sophisticated search techniques, whereas the second one is based on constraining the search space by performing a carefully designed pre-processing step. 
\paragraph{Using More Sophisticated Search}
One way of addressing the potential shortcomings of greedy structure selection is by using a more sophisticated search algorithm $\mathbb{A}$. For example, Biba et al.~\cite{biba:ilp08} used iterated local search as a way of avoiding local maxima when performing discriminative structure learning for MLNs. Iterative local search techniques \cite{lorenco:bookchapter03} alternate between two types of search steps, either moving towards a locally optimal solution, or perturbing the current solution in order to escape from local optima. 

An alternative approach is to search for structures of increasing complexity, at each stage using the structures found at the previous stage to constrain the search space. Such a strategy was employed by Khosravi et al.~\cite{khosravi:aaai10} for learning MLN structure in domains that contain many descriptive attributes. Their approach, \newstuff{which is similar to the technique employed to constrain the search space in PRMs \cite{friedman:ijcai99} described below,} distinguishes between two types of tables -- attribute tables that describe a single entity type, and relationship tables that describe relationships between entities. The algorithm, called MBN, then proceeds in three stages. In the first stage dependencies local to attribute tables are learned. In the second stage, dependencies over a join of an attribute table and a relationship table are learned, but the search space is constrained by requiring that all dependencies local to the attribute table found in the first stage remain the same. Finally, in the third stage dependencies over a join of two relationship tables, joined with relevant attribute tables, are learned, and the search space is similarly constrained. An orthogonal characteristic of MBN is that, although the goal is to learn an undirected SRL model,  dependencies are learned using a Bayesian network learner. The directed structures are then converted to undirected ones by ``moralizing'' the graphs. The advantage of this approach is that structure learning in directed models is significantly faster that structure learning in undirected models due to the decomposability of the score, which allows it to be updated locally, only in parts of the structure that have been modified, and thus scoring of candidate structures is more efficient.
\paragraph{Constraining the Search Space}
A second group of solutions is based on constraining the search space over structures, typically by performing a pre-processing step that, roughly speaking, finds some more promising regions of the space.

One approach, used for PRM learning, is to constrain the set of potential parents of each par-RV $X$~\cite{friedman:ijcai99}. This algorithm proceeds in stages, in each stage $k$, forming the set of potential parents of $X$ as those par-RVs that can be reached from $X$ through a chain of relations of length at most $k$. Structure learning at stage $k$ is then constrained to search only over those potential parent sets. Thus, so far, this algorithm is similar to techniques such as MBN described above. However, the algorithm further constrains potential parent candidates by requiring that they ``add value'' beyond what is already captured in the currently learned set of parents. More specifically, the set of potential parents of par-RV $X$ at stage $k$ consists of the parents in the learned structure from stage $k-1$, and any par-RVs reachable through relation chains of length at most $k$ that lead to a higher value in a specially designed score measure. This algorithm directly ports scoring functions that were developed for an analogous learning technique for Bayesian networks \cite{friedman:uai99}.

A series of algorithms in this group have been developed for learning MLN structure. The first in the series was BUSL~\cite{mihalkova:icml07}. BUSL is based on the observation that, once an MLN is instantiated into a Markov network, the instantiations of each clause of the MLN define a set of identically structured cliques in the Markov network. BUSL inverts this process of instantiation and constrains the search space by first inducing lifted templates for such cliques by learning a ``Markov network template,'' an undirected graph of dependencies whose nodes are not ordinary variables but par-RVs. Then clause search is constrained to the cliques of this Markov network template. Markov network templates are learned by constructing, from the perspective of each predicate, a table in which there is a row for each possible instantiation of the predicate and a column for possible par-RVs, with the value of a cell $i,j$ being set to 1 if the data contains a true instantiation of the $j$'th par-RV such that variable substitutions are consistent with the $i$'th predicate instantiation. The Markov network template is learned from this table by any Markov network learner.  

\begin{figure}[t]
\begin{center}
\includegraphics[scale=0.3]{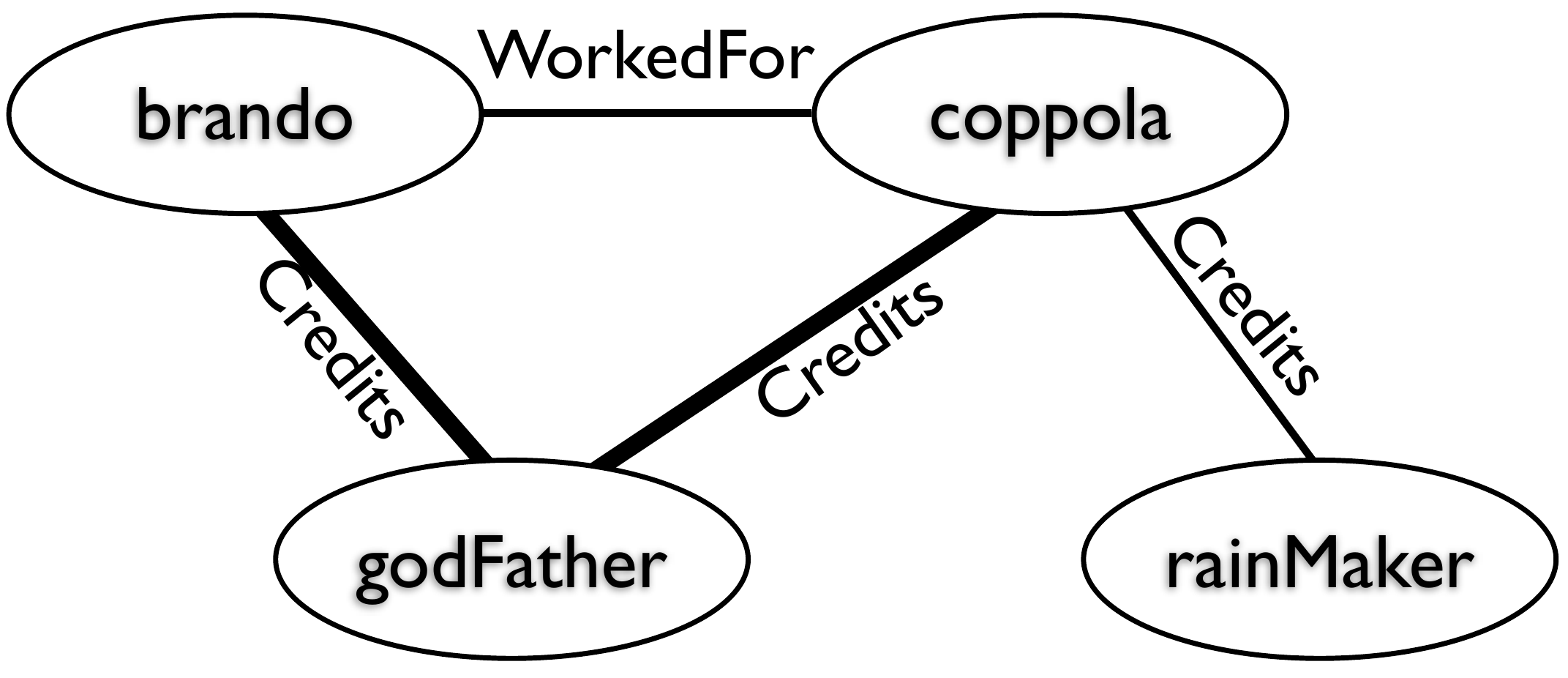}
\caption{Example of relational pathfinding.}
\label{fig:pathfinding}
\end{center}
\end{figure}

A further MLN learner that is based on constraining the search space is the LHL algorithm \cite{kok:icml09}. LHL limits the set of clause candidates that are considered by using relational pathfinding~\cite{richards:aaai92} to focus on more promising ones. Developed in the ILP community, relational pathfinding~\cite{richards:aaai92} searches for clauses by tracing paths across the true instantiations of relations in the data. Figure~\ref{fig:pathfinding} gives an example in which the clause $\mathtt{Credits(A,B)} \wedge \mathtt{Credits(A,C)} \Rightarrow \mathtt{WorkedFor(A,C)}$ is learned by tracing the thick-lined path between $\mathtt{brando}$ and $\mathtt{coppola}$ and variablizing appropriately. However, because in real-world relational domains the search space over relational paths may be very large, a crucial aspect of LHL is that it does not perform relational pathfinding over the original relational graph of the data but over a ``lifted hypergraph,'' which is formed by clustering the entities in the domain via an agglomerative clustering procedure, itself implemented as an MLN. Intuitively, entities are clustered together if they tend to participate in the same kinds of relations with entities from other clusters. Structure search is then limited only to clauses that can be derived as relational paths in the lifted hypergraph.

 Kok and Domingos~\cite{kok:icml10} have proposed constraining the search space by identifying ``structural motifs,'' which capture commonly occurring patterns among densely connected entities in the domain. The resulting algorithm, called LSM, proceeds by first identifying motifs and then searching for clauses by performing relational pathfinding within them. To discover motifs, LSM starts from an entity $i$ in the relational graph and performs a series of random walks. Entities that are reachable within a thresholded hitting time and the hyperedges among them are included in the motif and the paths via which they are reachable from $i$ are recorded. Next, the entities included in the motif are clustered by their hitting times into groups of potential symmetrical nodes. The nodes within each group are then further clustered in an agglomerative manner by the similarity of distributions over paths via which they are reachable from $i$. This process results in a lifted hypergraph, analogous to the one produced by LHL; however, whereas in LHL nodes were clustered based on their close neighborhood in the relational graph, here they are clustered based on their longer-range connections to other nodes. Motifs are extracted from the lifted hypergraphs via depth-first search.
\subsubsection{Structure Revision and Transfer Learning}
Our discussion so far has focused on learning structure from scratch. While approaches based on greedy search, such as Algorithm~\ref{alg:genericSL}, can be easily adapted to perform revision by starting learning from a given structure, some work in the area has also focused on approaches, specifically designed for structure revision and transfer learning. For example, Paes et al.~\cite{paes:ilp05} introduced an approach for revision of BLPs based on work in theory revision in the ILP community, where the goal is, given an initial theory, to minimally modify it such that it becomes consistent with a set of examples. The BLP revision algorithm follows the methodology of the FORTE theory revision system \cite{richards:mlj95}, first generating revision points in places where the given set of rules fails and next focusing the search for revisions to ones that could address the discovered revision points. The FORTE methodology was also followed in TAMAR, an MLN transfer learning system~\cite{mihalkova:aaai07}, which generates revision points on MLN clauses by performing inference and observing the ways in which the given clauses fail. TAMAR was designed for transfer learning, e.g.,\cite{icmlTLWkshop:06}, where the goal is to first map, or translate, the given structure from the representation of a source domain to that of a target and then to revise it. Thus, in addition to the revision module, it also contains a mapping module, which discovers the best mapping of the source predicates to the target ones. The problem of mapping a source structure to a target domain was also considered in the constrained setting where data in the target domain is extremely scarce~\cite{mihalkova:ijcai09}. Rather than taking a structure learned specifically for a source domain and trying to adapt it to a target domain of interest, an alternative approach to transfer learning is to extract general knowledge in the source domain that can then be applied to a variety of target domains. This is the approach taken in DTM~\cite{davis:icml09}, which uses the source data to learn general clique templates expressed as second-order Markov logic clauses, i.e. with quantification both over the predicates and the variables. During this step, care is taken to ensure that the learned clique templates capture general regularities and are not likely to be specific to the source domain. Then, in the target domain DTM allows for several possible mechanisms for using the clique templates to provide declarative bias. 
\subsubsection{Learning Causal Models}
An important structure learning problem is inducing causal models from relational data. This problem has been recently addressed by Maier et al.~\cite{maier:aaai10} whose RPC algorithm works with directed SRL models, which allow causal effects to be encoded in the directionality of the links. RPC extends its propositional analog, the PC algorithm \cite{spirtes:book01}, which has been developed for learning causality in graphical models. It proceeds in two stages. In the first stage, skeleton identification is performed to uncover the structure of conditional independencies in the data. In the second stage, edge orientation takes place, and orientations consistent with the skeleton are considered. RPC ports the edge orientation rules of PC to the relational setting and also develops new edge orientation rules that are specific to relational domains, in particular to model the uncertainty over whether a link exists between entities.
\section{Conclusion}
This article has presented a survey of work on lifted graphical models.   We have reviewed a general form for a lifted graphical model, a par-factor graph, and shown how a number of existing statistical relational representations map to this formalism.  We have discussed inference algorithms, including lifted inference algorithms, that efficiently compute the answers to probabilistic queries.  We have also reviewed work in learning lifted graphical models from data.  It is our belief that the need for statistical relational models (whether it goes by that name or another) will grow in the coming decades, as we are inundated with data which is a mix of structured and unstructured, with entities and relations extracted in a noisy manner from text, and with the need to reason effectively with this data.  We hope that this synthesis of ideas from many different research groups will provide an accessible starting point for new researchers in this expanding field.
\section*{Acknowledgement}
We would like to thank Galileo Namata and Theodoros Rekatsinas for their comments on earlier versions of this paper. L. Mihalkova is supported by a CI fellowship under NSF Grant \# 0937060 to the Computing Research Association. L. Getoor is supported by NSF Grants \# IIS0746930 and \# CCF0937094. Any opinions, findings, and conclusions or recommendations expressed in this material are those of the authors and do not necessarily reflect the views of the NSF or the CRA. 

\bibliographystyle{alpha}
\bibliography{mybib}
\end{document}